# On the Construction of the Inclusion Boundary Neighbourhood for Markov Equivalence Classes of Bayesian Network Structures

**Vincent Auvray**
Electrical Engineering and
Computer Science Dept.
University of Liège
auvray@montefiore.ulg.ac.be

**Louis Wehenkel**
Electrical Engineering and
Computer Science Dept.
University of Liège
Louis.Wehenkel@ulg.ac.be

## Abstract

The problem of learning Markov equivalence classes of Bayesian network structures may be solved by searching for the maximum of a scoring metric in a space of these classes. This paper deals with the definition and analysis of one such search space. We use a theoretically motivated neighbourhood, the inclusion boundary, and represent equivalence classes by essential graphs. We show that this search space is connected and that the score of the neighbours can be evaluated incrementally. We devise a practical way of building this neighbourhood for an essential graph that is purely graphical and does not explicitely refer to the underlying independences. We find that its size can be intractable, depending on the complexity of the essential graph of the equivalence class. The emphasis is put on the potential use of this space with greedy hill-climbing search.

## 1 INTRODUCTION

Learning Bayesian network structures is often formulated as a discrete optimization problem: the search for an acyclic structure maximizing a given scoring metric.

If we do not give any causal semantics to the arrow, we may consider that two Bayesian network structures are *(distribution) equivalent* if they can be used to represent the same set of probability distributions. Moreover, common scoring metrics assign the same value to equivalent structures and are thus also said *equivalent*. The ignorance of these facts may degrade the performance of greedy learning algorithms, but taken into account, they can also improve it (see [Chickering, 2002a] and [Andersson *et al.*, 1999]). In order to do so, we may assign to each equivalence class the score of its elements and search for the best class. The definition of a search space of equivalence classes is not trivial and is the topic of this paper. Our space is characterized by the fact that the classes are represented by essential graphs and the use of a notion of neighbourhood already proposed for Bayesian network structures: the *inclusion boundary neighbourhood*. Although this new space can be used by learning algorithms with various search strategies, we put the emphasis on greedy hill-climbing search. As we will see, the size of this neighbourhood is very large in the worst case but our results can serve as a basis for sensible approximations and are interesting by themselves.

In section 2, we present material used subsequently and mostly relevant to equivalence issues. The inclusion boundary is formally defined in section 3. Section 4 covers the generation of the neighbourhood and how to score its elements incrementally. Section 5 gives early comments on the hypothetical application of the space with greedy hill-climbing search. We conclude in section 6.

As a reviewer pointed out, the recent paper [Chickering, 2002b] partly deals with the same topic and presents, by another approach, corroborating results.

## 2 PRELIMINARIES

This section reviews notions required for a precise understanding of the paper, settles our notations and presents some of our theorems. The reading of these latter theorems is rather tedious and can be delayed until they are actually used in section 4.

### 2.1 GRAPHICAL NOTIONS

A *graph* $G$ is a pair $G = (V_G, E_G)$, where $V_G$ is a finite set of vertices and $E_G$ is a subset of $(V_G \times V_G) \setminus \{(a,a) | a \in V_G\}$. $E_G$ defines the *structure* of $G$ in the following way:

- $G$ contains a *line* between $a$ and $b$ if $(a,b) \in E_G$ and $(b,a) \in E_G$, which is noted $a - b \in G$,

- $G$ contains an *arrow* from $a$ to $b$ if $(a,b) \in E_G$ and $(b,a) \notin E_G$, which is noted $a \rightarrow b \in G$,

- $G$ contains an *edge* between $a$ and $b$ if $a - b, a \to b$ or $b \to a \in G$, which is noted $a \cdots b \in G$.

A graph is *complete* if there is an edge between every pair of distinct vertices. In this paper, all graphs considered in the search space have the same set of vertices $V$. There is thus a one-to-one correspondence between graphs and structures. The set of parents $pa_G(x)$ of a vertex $x$ in a graph $G$ consists of the vertices $y$ such that $y \to x \in G$.

A *subgraph* $G'$ of a graph $G$ is a graph such that[1] $V_{G'} \subsetneq V_G$ and $E_{G'} \subseteq E_G \cap (V_{G'} \times V_{G'})$. The *induced subgraph* $G_A$, where $A \subseteq V_G$ is the subgraph of $G$ such that $V_{G_A} = A$ and $E_{G_A} = E_G \cap (A \times A)$. A set of vertices is *complete* in $G$ if the subgraph of $G$ it induces is complete.

A *path* is a sequence $x_0, \ldots, x_n$ of distinct vertices where $(x_i, x_{i+1}) \in E_G, i = 0, \ldots, n-1$. This path is undirected if $x_n, \ldots, x_0$ is also a path. The relation $\approx$ is an equivalence relation between vertices defined by $a \approx b \Leftrightarrow a = b$ or there exists an undirected path between $a$ and $b$. This relation partitions the set of vertices of a graph into equivalence classes. A *cycle* is a path with the modification that $x_0 = x_n$. A cycle is *directed* if $x_i \to x_{i+1} \in E_G$ for a least one $i \in \{0, \ldots, n-1\}$.

A *v-structure* $(h, \{t_1, t_2\})$ of $G$, where $h, t_1$ and $t_2$ are distinct vertices is a pair such that $t_1 \to h \in G, t_2 \to h \in G$ and $t_1 \cdots t_2 \notin G$. $V(G)$ denotes the set of v-structures of $G$. An arrow $p \to q \in G$ is *protected* in $G$ if $pa_G(p) \neq pa_G(q) \setminus \{p\}$. An arrow $p \to q \in G$ is *strongly protected* if $G$ induces at least one of the subgraphs of figure 1.

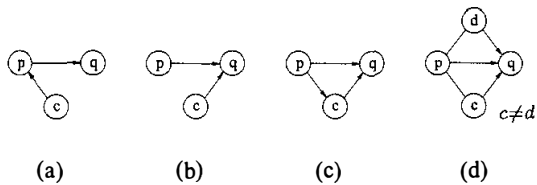

Figure 1: Strongly Protected Arrow $p \to q$

According to their properties, graphs may be classified in directed, undirected and chain graphs. A *directed graph* $D = (V_D, E_D)$ is a graph without any line. An *acyclic directed graph* or *DAG* is a directed graph that contains no (directed) cycle. The set of all DAGs defined with the same set of vertices $V$ is noted $\mathcal{D}$. Obviously, an arrow $a \to b \in D$ is protected if $D$ induces at least one of the subgraphs (a), (b) or (c) of figure 1.

An *undirected graph* $U = (V_U, E_U)$ is a graph without any arrow. The *skeleton* $S(G)$ of a graph $G$ is the undirected graph resulting from ignoring the orientation of the arrows in $G$. An undirected graph is *chordal* if every (undirected) cycle of length $\geq 4$ has a chord, i.e. a line between two non-consecutive vertices of the cycle. A *perfect ordering* of an undirected graph $U$ is a total ordering of $V_U$ such that the directed graph $D$ obtained by directing every line $a - b \in U$ from $a$ to $b$ if $a$ precedes $b$ in the ordering is acyclic and contains no v-structure. $D$ is called a *perfect directed version* of $U$. The following theorem holds true (see e.g. [Cowell *et al.*, 1999]).

**Theorem 2.1** *An undirected graph is chordal if, and only if, it admits at least one perfect ordering.*

*Maximum cardinality search* (*MCS*) is an algorithm that checks if an undirected graph is chordal and, if so, provides a perfect ordering. A description of MCS can be found in [Cowell *et al.*, 1999]. Let us just note that MCS can be used to immediately and *constructively* prove lemma 2.2.

**Lemma 2.2** *Given a chordal undirected graph $U$ and a non-empty set of vertices $A \subseteq V_U$ inducing a complete subgraph, any permutation of $A$ is the beginning of a perfect ordering of $U$.*

For example, let $a$ and $b$ be a pair of adjacent vertices of $U$. We deduce from the application of the lemma 2.2 with $A = \{a\}$ that there exists a perfect directed version $D$ of $U$ such that $a \to b \in D$.

The next lemma is used in theorem 2.7.

**Lemma 2.3** *The removal of $a - b$ from a chordal undirected graph $U$ that does not induce the subgraph of figure 9 for any $h \in V_U$ produces a chordal undirected graph.*

**Proof.** Let us prove that if $a - b$ is a chord for a cycle of length $m \geq 4$, then $U$ induces the subgraph of figure 9. $a - b$ divides that cycle into two sub-cycles. Let $x$ be one of them. On the one hand, the existence of a cycle $c$ of length 3 containing $a - b$ implies that $U$ induces the subgraph of figure 9. On the other hand, if there exists a cycle $c$ of length $n \geq 4$ containing $a - b$ then there exists a cycle $c'$ containing $a - b$ and of length $n'$ such that $3 \leq n' \leq n - 1$. Such a $c'$ can be obtained as follows. By the chordality of $U$, $c$ has (at least) one chord. That chord divides $c$ into two sub-cycles. $c'$ can be chosen as the sub-cycle containing $a - b$. By an inductive reasoning starting with $c = x$ and ending with $n = 3$, we conclude that $U$ induces the subgraph of figure 9. ∎

A *consistent extension* of a graph $G$ is a DAG $D$ such that $D$ has the same skeleton as $G$, the same set of v-structures and every arrow of $G$ is present in $D$. Dor and Tarsi (see [Dor and Tarsi, 1992]) found an algorithm to check if a graph possesses a consistent extension and, if so, to find one.

---

[1] Throughout this paper, $A \subset B$ means that $A$ is a proper subset of $B$, while $A \subseteq B$ means that $A = B$ or $A \subset B$.

A *chain graph* $C$ is a graph without any directed cycle. DAGs and undirected graphs are special cases of chain graphs. The set of *chain components* of a chain graph $C$ is the set of equivalence classes of vertices induced by the relation $\approx$ in $C$. Each subgraph of $C$ induced by a chain component is undirected, because otherwise $C$ would contain a directed cycle.

## 2.2  BAYESIAN NETWORKS

Graphs are sometimes used to represent sets of conditional independences between random variables.

A *Bayesian network* $B$ for a set of random variables $X = \{x_1, \ldots, x_n\}$ is a pair $(D, \Theta)$, where $D$ is a DAG defined on a set of vertices in one-to-one correspondence with $X$ and $\Theta = \{\theta_1, \ldots, \theta_n\}$ is a set of parameters such that each $\theta_i$ defines a conditional probability distribution $P(x_i | pa_D(x_i))$. Such a Bayesian network represents the probability distribution $P(X)$ defined as $P(X) = \prod_{i=1}^{n} P(x_i | pa_D(x_i))$.

Let us define $I(D)$ as the set of conditional independences $U \perp V | W$[2] such that $W$ d-separates[3] $U$ and $V$ in $D$. One can show that the independences of $I(D)$ are verified in $P(X)$. Conversely, if a probability distribution $P(X)$ verifies the independences of a set $I(D)$, then $P(X)$ can be decomposed in a product $\prod_{i=1}^{n} P(x_i | pa_D(x_i))$ and is thus representable by a Bayesian network defined on $D$.

## 2.3  EQUIVALENCE OF DAGS

Two DAGs $K, L \in \mathcal{D}$ (or their structure) are *independence* (or *Markov*) *equivalent* if $I(K) = I(L)$. This relation induces equivalence classes in $\mathcal{D}$.

Distribution equivalence implies independence equivalence, but the converse is not true in general. The subsequent developments are all based on independence equivalence, even if not explicitely mentioned. To use them in our learning problem, we place this paper in any context where independence and distribution equivalences are logically equivalent.

Verma and Pearl (1990) derived the following theorem.

**Theorem 2.4** *Two DAGs are equivalent if, and only if, they have the same skeleton and the same set of v-structures.*

An equivalence class is thus characterized by a skeleton and a set of v-structures.

The *essential graph*[4] $E = (V_E, E_E)$ of an equivalence class noted $[E]$ is defined by $V_E = V_D$ for any $D \in [E]$ and $E_E = \cup_{D \in [E]} E_D$. Let us make a few comments about $E$. $E$ has the same skeleton as the DAGs of $[E]$. $E$ has the arrow $a \to b$ if, and only if, every DAG of $[E]$ also has it. Similarly, $a - b \in E$ if, and only if, there exist two DAGs of $[E]$ such that one has the arrow $a \to b$ and the other has $b \to a$. For example, we deduce from theorem 2.4 that $E$ and the DAGs of $[E]$ have the same skeleton and set of v-structures.

---
[2]The notation $U \perp V | W$ means that the sets of variables $U$ and $V$ are independent given the set of variables $W$.
[3]See [Cowell *et al.*, 1999] or [Pearl, 1988] for a definition of d-separation and all the details.
[4]Essential graphs are also called completed partially directed acyclic graphs.

Let $D^*$ denote the essential graph of the equivalence class containing the DAG $D$. The following theorem ensures that the essential graph $E$ can be used as a representation of $[E]$ (see [Andersson *et al.*, 1999]).

**Theorem 2.5** *Let $K, L$ be two DAGs. $I(K) = I(L)$ if, and only if, $K^* = L^*$.*

The set of (conditional) independences $I(E)$ represented by an essential graph $E$ is defined as the set $I(D)$ of (conditional) independences of any $D \in [E]$.

Essential graphs are characterized by a theorem of Andersson (see [Andersson *et al.*, 1999]).

**Theorem 2.6** *A graph $G$ is an essential graph, i.e. $G = D^*$ for some DAG $D$ if, and only if, $G$ satisfies the following conditions:*

- *$G$ is a chain graph;*
- *for every chain component $\tau$ of $G$, $G_\tau$ is chordal;*
- *$G$ does not induce the subgraph $a \to b - c$;*
- *every arrow of $G$ is strongly protected in $G$.*

Let $\mathcal{E}$ denote the set of essential graphs defined on the set of vertices $V$.

We proved the following theorem, used in section 4.2.

**Theorem 2.7** *Let $E$ be an essential graph such that $a - b \in E$ and $E$ does not induce the subgraph of figure 9 for any $h \in V_E$. The graph $G$ obtained by removing $a - b$ from $E$ is essential.*

**Proof.** Obviously, $G$ is a chain graph and does not induce $v_1 \to v_2 - v_3$. Let $S$ be an induced subgraph of $E$ strongly protecting an arrow $p \to q$. If $S$ is of the type of figure 1(d) and $a - b \in S$, then $G$ induces a subgraph $S'$ of the type of figure 1(b). Otherwise, $S$ is also an induced subgraph of $G$. Every arrow of $G$ is thus strongly protected in $G$. Let $\tau$ be a chain component of $E$. If $a - b \in E_\tau$ then by lemma 2.3, $G_\tau$ is chordal. Otherwise, $G_\tau = E_\tau$. Hence, every subgraph of $G$ induced by one of its chain components is chordal. By theorem 2.6, $G$ is an essential graph. ∎

The essential graph $D^*$ can be obtained from $D$ by the following algorithm and theorem from [Andersson et al., 1999].

**Algorithm 2.1** Let $G_0$ be a graph. For $i \geq 1$, convert every arrow $a \to b \in G_{i-1}$ that is not strongly protected in $G_{i-1}$ into a line, obtaining a graph $G_i$. Stop as soon as $G_k = G_{k+1}(k \geq 0)$ and return $G_k$.

**Theorem 2.8** If $G_0 = D$, then algorithm 2.1 returns $D^*$.

Note that other algorithms exist, but we extend this one by theorem 2.10. Let us first introduce a lemma and some notation. Define $Q(G)$ as the set of arrows of the graph $G$ that are strongly protected in $G$ only by one or more subgraphs of the type of figure 1(d).

**Lemma 2.9** Let $S$ and $L$ be graphs such that

(i) every arrow of $L$ is strongly protected in $L$,

(ii) every arrow of $L$ is present in $S$,

(iii) for each $a \to b \in Q(L)$, $S$ induces one of the subgraphs of figure 2.

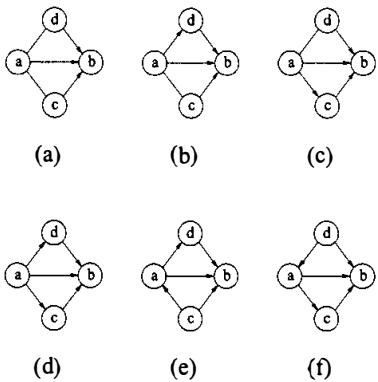

Figure 2: Induced Subgraphs

Let $S'$ be the graph obtained from $S$ by converting every non strongly protected arrow into a line. The graph $S'$ satisfies the above hypotheses concerning $S$.

**Proof.** Let $a \to b$ be an arrow of $L$. Suppose first that $a \to b \in Q(L)$. Let us consider the induced subgraphs of figure 2. In each case, the arrows $c \to b$ and $d \to b$ strongly protect one another by a subgraph of the type of figure 1(b) induced on $\{b, c, d\}$. In case 2(a), $a \to b$ is strongly protected by an induced subgraph of the type of figure 1(d), while in the other cases, that arrow is strongly protected by (at least) one induced subgraph of the type of figure 1(c). Moreover, in case 2(e), $a \to d$ is strongly protected by a subgraph of the type of figure 1(a) on $\{a, c, d\}$. Similarly, in case 2(f), $a \to c$ is strongly protected. By construction, $S'$ thus induces one of the subgraphs of figure 2, and in particular $a \to b \in S'$. Suppose now that $a \to b \notin Q(L)$. By hypothesis (i), $a \to b$ is strongly protected in $L$ by (at least) one induced subgraph of the type of figure 1(a), 1(b) or 1(c). Because every arrow of $L$ is present in $S$, $a \to b$ is also strongly protected in $S$ and thus $a \to b \in S'$. ∎

**Theorem 2.10** Let $D$ be a DAG and $G_0$ a graph such that

(i) $V_{G_0} = V_D$,

(ii) every arrow of $G_0$ is present in $D$,

(iii) every arrow of $D^*$ is present in $G_0$,

(iv) for each $a \to b \in Q(D^*)$, $G_0$ does not induce the subgraph of figure 3 for any $c, d \in V_{G_0}$.

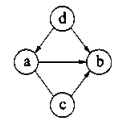

Figure 3: Forbidden Subgraph

*Algorithm 2.1 applied to $G_0$ returns $D^*$.*

**Proof.** Let $G_0, \ldots, G_k$ be the sequence of graphs produced by algorithm 2.1. Obviously, $S(G_k) = S(D^*)$. Let us show that $G_k$ and $D^*$ have the same arrows and thus $G_k = D^*$. On the one hand, let $a \to b$ be an arrow of $Q(D^*)$. $D^*$ thus induces a subgraph of the type of figure 1(d) to strongly protect $a \to b$. By hypotheses (iii) and (iv), $G_0$ induces one of the subgraphs of figure 2. By an inductive application of lemma 2.9 beginning with $S = G_0$ and $L = D^*$, we deduce that the arrows of $D^*$ are present in $G_k$. On the other hand, note that by hypothesis (ii) and the description of algorithm 2.1, every arrow of $G_k$ is present in $D$. If $a \to b \in Q(G_k)$, $D$ thus induces a subgraph of the type of figure 2(d), 2(e) or 2(f). We deduce from theorem 2.8 and an inductive application of lemma 2.9 beginning with $S = D$ and $L = G_k$ that the arrows of $G_k$ are present in $D^*$. ∎

The first three hypotheses of this theorem are, for example, satisfied if $G_0$ is obtained from $D$ by converting some arrows $a \to b \in D$ such that $a - b \in D^*$ into lines, or if $G_0$ is such that $V_{G_0} = V_D$ and $E_{G_0} = \cup_{G \in X} E_G$, where $X \subseteq [D^*]$.

Conversely, every DAG $D \in [E]$ can be recovered from $E$ by theorem 2.11 (see [Andersson et al., 1999]).

**Theorem 2.11** $D \in [E]$ if, and only if, $D$ is obtained from $E$ by orienting the lines of every undirected (chordal) subgraph induced by a chain component of $E$ according to a perfect ordering.

As expected, by definition of the perfection of an ordering, the orientation of the lines does not introduce any new v-structure. Besides, one can see that the elements of $[E]$ are the consistent extensions of $E$. Dor and Tarsi's algorithm applied to $E$ thus returns a $D \in [E]$.

## 2.4 SCORING METRICS

A scoring metric *score* for DAGs is *decomposable* if it can be written as a sum (or product) of functions[5] of only one vertex and its parents, i.e.

$$score(D) = \sum_{x \in V} f(x, pa_D(x))$$

A scoring metric for DAGs is *equivalent* if it assigns the same value to equivalent DAGs. In this paper, this property is supposed to hold, as for example with the well-known BDe score. In such a case, the score of an equivalence class (or its essential graph) is defined as the score of (any of) its elements.

## 3 DEFINITION OF THE INCLUSION BOUNDARY NEIGHBOURHOOD

The neighbourhood of a Bayesian network structure is often defined in terms of operations performed on that structure, such as the addition, removal or reversal of an arrow. For example, the graphs of figure 4 are typically neighbours. This kind of neighbourhood is constructed very simply and efficiently. Furthermore, if a decomposable scoring metric is used, the score of the neighbours of a structure can be calculated incrementally, i.e. with just a few evaluations of $f$.

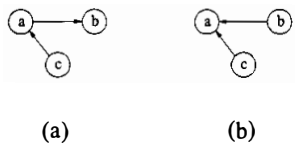

Figure 4: Adjacent DAGs

The same idea is applicable to search spaces of essential graphs, with operators such as the addition of an arrow, a line or a v-structure, the reversal of an arrow,... However, the situation is complicated by the constraints on essential graphs: the graph modified by an operator must satisfy the conditions of theorem 2.6. The recent paper [Chickering, 2002a] shows that these problems can be overcome by carefully choosing the operators so as to finally get an efficient algorithm, and in particular keep an incremental evaluation of the scoring metric.

In these latter two cases, the neighbourhood is defined by modifications performed on the graph, without any reference to the independences represented. Instead, it may be defined as its inclusion boundary. Let $\mathcal{G}$ be a set of graphs representing (conditional) independences. A graph $G' \in \mathcal{G}$ belongs to the *inclusion boundary with respect to* $\mathcal{G}$ of $G \in \mathcal{G}$ if $G' \neq G$ and one of the following mutually exclusive conditions is satisfied:

(i) $I(G') = I(G)$,

(ii) $I(G) \subset I(G')$ and there is no $G'' \in \mathcal{G}$ verifying $I(G) \subset I(G'') \subset I(G')$,

(iii) $I(G') \subset I(G)$ and there is no $G'' \in \mathcal{G}$ verifying $I(G') \subset I(G'') \subset I(G)$.

This idea has already been used in [Kočka and Castelo, 2001] with $\mathcal{G} = \mathcal{D}$, i.e. with Bayesian network structures. The DAGs of figure 4 are then no longer neighbours. We transpose this idea to define a space based on equivalence classes represented by essential graphs, i.e. $\mathcal{G} = \mathcal{E}$. In this case, the first condition is never satisfied. For a particular $E \in \mathcal{E}$ the set of essential graphs defined by (ii) and (iii) are respectively noted $N^+(E)$ and $N^-(E)$. By definition, these sets never intersect. Note that if $M \in N^+(N)$, then obviously $N \in N^-(M)$, and conversely.

Our search space is *connected* if, between any $M, N \in \mathcal{E}$, there exists a finite sequence of essential graphs $E_1, \ldots, E_l$, such that $E_1 = M$, $E_l = N$ and $E_{i+1}$ is a neighbour of $E_i$ for $i = 1, \ldots, l - 1$. This property is important for local search.

**Theorem 3.1** *The search space is connected.*

**Proof.** There exists an essential graph $U$ defined on the finite set of vertices $V = \{v_1, \ldots, v_n\}$ (i.e. $U \in \mathcal{E}$) and such that $I(U) = \emptyset$. Indeed, let $D$ be the DAG such that $pa_D(v_i) = \{v_1, \ldots, v_{i-1}\}$. We have $U = D^*$. For each $E \in \mathcal{E}$, the following facts hold. If $E \neq U$, $I(U) \subset I(E)$ and thus $N^-(E) \neq \emptyset$. For all $G \in N^-(E)$, $|I(G)| < |I(E)|$[6]. The set $I(E)$ is finite. Hence, there exists a finite sequence of essential graphs $E_1, \ldots, E_l$, such that $E_1 = E$, $E_l = U$ and $E_{i+1} \in N^-(E_i)$. By the symmetry of the neighbourhood, the sequence $E_l, \ldots, E_1$ is such that $E_i \in N^+(E_{i+1})$. For all $M, N \in \mathcal{E}$, there thus exists a sequence $M, \ldots, U, \ldots, N$ with the required properties. ∎

---

[5]The dependence of the metric on the data is not made explicit.

[6]$|X|$ denotes the cardinality of the set $X$.

Some questions still need an answer. Is the size of this neighbourhood tractable? Can the elements of the neighbourhood be generated efficiently and/or scored incrementally? The next section addresses them by explicitely building $N^+(E)$ and $N^-(E)$.

## 4 CONSTRUCTION OF THE INCLUSION BOUNDARY NEIGHBOURHOOD

The construction of $N(E) = N^+(E) \cup N^-(E)$ from the conditions (ii) and (iii) is not immediate. Moreover, given $G \in \mathcal{E}$, it is not trivial to check whether $G \in N(E)$ or not. These difficulties stem from the fact that the conditions are expressed through $I(E)$ instead of $E$'s graphical components.

The following lemma[7] simplifies the expression of the neighbourhood.

**Lemma 4.1**

$$N^+(E) = \{G \in \mathcal{E} | \exists K, L \in \mathcal{D} : K^* = G, L^* = E$$
$$\text{and } K \text{ is obtained from } L \text{ by the}$$
$$\text{removal of one arrow}\},$$

$$N^-(E) = \{G \in \mathcal{E} | \exists K, L \in \mathcal{D} : K^* = G, L^* = E$$
$$\text{and } K \text{ is obtained from } L \text{ by the}$$
$$\text{addition of one arrow}\}.$$

If a decomposable scoring metric is used, an important corollary is the possibility to evaluate incrementally the score of $E$'s neighbours from the score of $E$. Using the notations of the last lemma, for each $G \in N(E)$ we have

$$\Delta_G score = score(G) - score(E),$$
$$= score(K) - score(L),$$
$$= f(x, pa_K(x)) - f(x, pa_L(x)), \quad (1)$$

where $x$ is the destination of the arrow added or removed in $L$. We see that very little is sufficient to estimate the increment in score, in particular the complete knowledge of $G$ is not needed.

The expressions of lemma 4.1 suggest a practical way of building $N(E)$. Given a DAG $D$, let $\mathcal{R}(D)$ be the set of DAGs that can be constructed by removing or adding an arrow to $D$. Obviously, we have $N(E) = \cup_{D \in [E]} \cup_{M \in \mathcal{R}(D)} \{M^*\}$. However, this approach is redundant in the sense that the sets of the unions are not necessarily disjoint. For example, the DAGs of figures 5(a) and 5(b) belong to the equivalence class represented by the essential graph of figure 5(c). The removal of $a \to b$ from both of these produces DAGs of the same class, represented by figure 5(d).

---

[7]Our proof of this lemma uses *Meek's conjecture*, recently proved in [Chickering, 2002b], and can be obtained upon request to the first author.

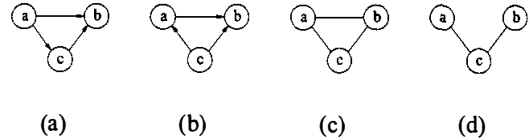

(a)  (b)  (c)  (d)

Figure 5: Removal Producing Equivalent DAGs

To circumvent this pitfall, we divide our task in two parts: the identification of the neighbours and, if necessary, their construction. Let $c : N(E) \to X$ be an injective function, i.e. such that $G_1 \neq G_2$ implies $c(G_1) \neq c(G_2)$. In this paper, $c$ is called a *characterization function* and $c(G)$ a *characterization of $G$*. A given $x \in X$ is *valid* if it characterizes a $G \in N(E)$, i.e. there exists a $G \in N(E)$ such that $c(G) = x$.

With a characterization function $c$, we may build $N(E)$ by first identifying the valid elements of $X$ and then, for each such element, obtaining the corresponding $G \in N(E)$. We will use two such functions: $c_1$ to build $N^-(E)$ and $c_2$ for $N^+(E)$. They are defined as follows. We deduce from theorem 2.4 that each $G \in N(E)$ is characterized by its skeleton $S(G)$ and set of v-structures $V(G)$. Let $T$ be the set of skeletons that are obtained from $S(E)$ by removing or adding a line. There is a one-to-one correspondence between $T$ and the set of unordered pairs of vertices such that each $t \in T$ is associated to the pair $\{a, b\}$ of vertices that are the endpoints of the line removed or added to obtain $t$ from $E$. We see from lemma 4.1 that, for each $G \in N(E)$, $S(G) \in T$.

Besides, the set $V(G)$ for $G \in N(E)$ can be decomposed as $(V(G) \setminus V(E)) \cup (V(E) \setminus (V(E) \setminus V(G)))$. By lemma 4.1, $V(G) \setminus V(E)$ and $V(E) \setminus V(G)$ are the sets of v-structures respectively created and destroyed by the addition or removal[8] of an arrow in a $D \in [E]$. We deduce from the following obvious lemma that $V(E) \setminus V(G)$ depends only on $S(G)$ (or the pair of vertices corresponding to it).

**Lemma 4.2** *If the execution of the operation corresponding to a given pair $\{a, b\}$ destroys a v-structure $v$ from a $D \in [E]$ then that operation destroys $v$ from every $G \in [E]$.*

Gathering these observations, we have a first characterization function $c_1 : N(E) \to X_1 : G \to c_1(G) = (g_1(G), g_2(G)) = (\{a, b\}, O)$, where $\{a, b\}$ are the vertices associated to $S(G)$ and $O = V(G) \setminus V(E)$. The previous discussion also leads to the validity condition of lemma 4.3.

**Lemma 4.3** *A pair $(\{a, b\}, O) \in X_1$ characterizes $G \in N(E)$ if, and only if, $\exists K, L \in \mathcal{D} : K^* = G, L^* = E, K$ is*

---

[8]This operation on DAGs mirrors the operation performed on $S(E)$ to obtain $S(G)$.

*obtained from L by performing the operation associated to* $\{a,b\}$ *and* $O = V(K) \setminus V(L)$.

Let $\mathcal{A}(v, \{a,b\}) \subseteq [E]$ denote the set of DAGs where the operation associated to $\{a,b\}$ creates the v-structure $v$ and let $R(\{a,b\})$ denote the set $\{v|\mathcal{A}(v, \{a,b\}) \neq \emptyset\}$. Obviously, if $(\{a,b\}, O)$ is valid then $O \subseteq R(\{a,b\})$.

We also use a slightly modified version of $c_1$, defined as follows. Given $\{a,b\}$, let $Y$ be the set $\cap_{O \in Z} O$ where $Z = \{O|(\{a,b\}, O) \in X_1$ is valid$\}$. In other words, $Y$ consists of the v-structures that are created in every DAG of $[E]$ by the operation. The function $c_2 : N(E) \to X_2 : G \to c_2(G) = (g_1(G), g_2(G) \setminus Y)$ is injective. Let $W(\{a,b\})$ denote the set $\{v|\mathcal{A}(v, \{a,b\}) \neq \emptyset$ and $\mathcal{A}(v, \{a,b\}) \neq [E]\}$. The validity condition for $c_2$ is given by the next lemma.

**Lemma 4.4** *A pair* $(\{a,b\}, O) \in X_2$ *characterizes* $G \in N(E)$ *if, and only if,* $O \subseteq W(\{a,b\})$ *and* $\exists K, L \in \mathcal{D} :$ $K^* = G, L^* = E, K$ *is obtained from $L$ by performing the operation associated to* $\{a,b\}$, $O \subseteq V(K) \setminus V(L)$ *and* $W(\{a,b\}) \setminus O \not\subseteq V(K) \setminus V(L)$.

In sections 4.1 to 4.3 we present our method to identify the valid characterizations and determine the corresponding neighbours and increments in score. As we will see, the method differs if there is an arrow between $a$ and $b$ in $E$, $a - b \in E$ or $a \cdots b \notin E$. Let $N_{ab}(E)$ be the subset of $N(E)$ such that its elements have the same skeleton, characterized by $\{a,b\}$. By analogy with the other type of neighbourhood cited in section 3, we define three pseudo-operators[9]: *removal of* $a \to b \in E$, *removal of* $a - b \in E$ and *addition of an edge between $a$ and $b$ to $E$* used in the corresponding situations and returning $N_{ab}(E)$. The construction of $N(E)$ can then proceed by enumerating the unordered pairs of vertices and, for each, calling the corresponding pseudo-operator.

We have the following theorem.

**Theorem 4.5** *For each* $\{a,b\}$, $N_{ab}(E)$ *is non-empty.*

**Proof.** Let $D$ be a DAG of $[E]$. If $a \cdots b \in D$, then the graph $D'$ obtained by removing that edge from $D$ is obviously a DAG. If $a \cdots b \notin D$ then there exists a DAG $D'$ obtained from $D$ by adding an arrow between $a$ and $b$. Indeed, suppose that the addition of $a \to b$ creates the cycle $a, b, v_{i_1}, \ldots, v_{i_k}, a$ and the addition of $b \to a$ creates the cycle $b, a, v_{i_{k+1}}, \ldots, v_{i_{k+l}}, b$. $D$ would possess the cycle $b, v_{i_1}, \ldots, v_{i_k}, a, v_{i_{k+1}}, \ldots, v_{i_{k+l}}, b$ and would not be a DAG. By lemma 4.1, $D'^* \in N_{ab}(E)$. ∎

## 4.1 REMOVAL OF AN ARROW $a \to b \in E$

In the context of application of this pseudo-operator, $\{a,b\}$ is fixed and the associated operation is the removal of $a \to b$. We use the characterization function $c_2$.

The lemma 4.6 obviously holds true.

**Lemma 4.6** *The removal of an arrow $a \to b$ from a DAG creates the v-structure* $(h, \{t_1, t_2\})$ *if, and only if,* $\{t_1, t_2\} = \{a,b\}$ *and the DAG induces the subgraph of figure 6.*

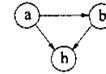

Figure 6: Creation of a V-Structure

The set $W(\{a,b\})$ is easily identified graphically from $E$ with the following theorem.

**Theorem 4.7** $(h, \{t_1, t_2\}) \in W^{[10]}$ *if, and only if,* $\{t_1, t_2\} = \{a,b\}$ *and $E$ induces the subgraph of figure 7.*

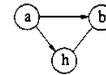

Figure 7: Induced Subgraph of $E$

**Proof.** If we remind the meaning of a line of an essential graph, the sufficient part is trivial[11]. By lemma 4.6 and the definition of $W$, $\{t_1, t_2\} = \{a,b\}$ and there exists a $D \in [E]$ inducing the subgraph of figure 6. There also exists a $K \in [E]$ where the removal of $a \to b$ does not create $(h, \{a,b\})$. Such a $K$ must have $a \to b$ and the same skeleton as $D$. It thus induces the acyclic subgraph 8(a) or 8(b). By the acyclicity of $E$, $K$ must induce 8(b) and

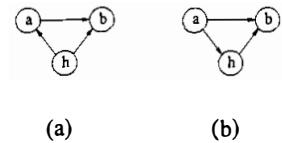

(a)                (b)

Figure 8: Induced Subgraphs of $K$

$a \to h \in E$. $E$ thus induces the subgraph of figure 7. ∎

The valid characterizations $O$ are obviously subsets of $W$ and can be obtained with the following theorem.

---

[9]These are not operators in the usual sense because they return a set of states instead of a single one.

[10]The dependence on $\{a,b\}$ is made implicit for brevity.

[11]The sufficient part is not used in theorem 4.8.

**Theorem 4.8** *$O$ is valid if, and only if, $O \subseteq W$ and the set $C = \{h|(h, \{a,b\}) \in W \setminus O\}$ is complete in $E$.*

**Proof.** By lemma 4.4, $O \subseteq W$ is valid if, and only if, there exists a $D \in [E]$ such that $b \rightarrow h \in D$ for $h \in \{h|(h,\{a,b\}) \in O\}$ and $h \rightarrow b \in D$ for $h \in C$. The existence of such a $D$ is checked with theorem 2.11. Let $\tau$ be the chain component of $E$ containing $\{h|(h,\{a,b\}) \in W\}$. The constraints on the orientation of the lines of $E$ to obtain $D$ are only related to $E_\tau$. Each subgraph of $E$ induced by another chain component can thus be directed according to a perfect ordering[12] independently. Hence, there exists such a $D$ if, and only if, there exists a perfect ordering of $E_\tau$ leading to the required arrows. On the one hand, let $o$ be such a perfect ordering. The perfect directed version $H$ of $E_\tau$ has no v-structure and the arrows $h \rightarrow b$ for $h \in C$. Any vertices $h_i, h_j \in C$ must be adjacent in $H$, because otherwise $H$ would possess the v-structure $(b, \{h_i, h_j\})$. We thus have $h_i - h_j \in E$. On the other hand, suppose that $C$ is complete. $C \cup \{b\}$ is then also complete. By lemma 2.2, for any permutation $h_1, \ldots, h_k$ of $C$, $h_1, \ldots, h_k, b$ is the beginning of a perfect ordering $o$. Such an ordering leads to the required arrows. ∎

This theorem has an immediate corollary.

**Corollary 4.9** *There is a one-to-one mapping between $N_{ab}(E)$ and the complete subsets of $\{h|(h,\{a,b\}) \in W\}$.*

Suppose that $O$ characterizes $E' \in N_{ab}(E)$. Let us discuss the construction of $E'$. We use the notations of the previous theorem. Let $D$ be a DAG obtained from E by (i) removing $a \rightarrow b$, (ii) directing the lines of $E_\tau$ according to $o$ and (iii), for each subgraph $E_\alpha$ of $E$ induced by another chain component, directing its lines according to a perfect ordering. From the proof of theorem 4.8, we see that the set $\mathcal{B}$ of these DAGs is a subset of $[E']$. The graph $G$ such that $E_G = \cup_{D \in \mathcal{B}} E_D$ can clearly be constructed from $E$ by performing the steps (i) and (ii). Moreover, by symmetry of $o$, we know that $E'_C$ is undirected. Let us undirect the arrows of $G$ that are present in $G_C$. We have the following result.

**Theorem 4.10** *Algorithm 2.1 applied to $G$ returns $E'$.*

**Proof.** Let us show that $G$ satisfies the hypotheses of theorem 2.10. Obviously, $G$ satisfies the first three conditions. Let us show that $G$ does not induce $p \rightarrow q - r$ and thus satisfies the last hypothesis. Suppose that $G$ induces $p \rightarrow q-r$. If $p \rightarrow q \in E$, then by construction of $G$, $E$ also induces $p \rightarrow q-r$, which is impossible since $E$ is an essential graph. Otherwise, using the notations of theorem 4.8, $q, r \in C$ and $p \in \tau \setminus C$. But the arrows of $G$ are directed according to an ordering beginning with a permutation of $C$. Thus, there can not be an arrow from a vertex of $\tau \setminus C$

---
[12]The existence of such an ordering is guaranteed by theorems 2.6 and 2.1.

to one of $C$. ∎

In a sense, $E'$ can thus be constructed incrementally from $E$.

The increment in score is easily evaluated with formula (1), yielding:

$$\Delta_G score = f(b, (pa_E(b) \setminus \{a\}) \cup C) - f(b, pa_E(b) \cup C)$$

### 4.2 REMOVAL OF A LINE $a - b \in E$

The operation associated to $\{a, b\}$ is the removal of the arrow between $a$ and $b$. We use $c_2$. The set $W(\{a,b\})$ can be identified graphically by the following theorem, the proof of which is very similar to that of theorem 4.7.

**Theorem 4.11** *$(h, \{t_1, t_2\}) \in W$ if, and only if, $\{t_1, t_2\} = \{a, b\}$ and $E$ induces the subgraph of figure 9.*

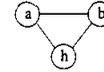

Figure 9: Induced Subgraphs of $E$

The valid characterizations $O$ are subsets of $W$ and are found with theorem 4.12, whose terms are identical to those of theorem 4.8.

**Theorem 4.12** *$O$ is valid if, and only if, $O \subseteq W$ and the set $C = \{h|(h,\{a,b\}) \in W \setminus O\}$ is complete in $E$.*

**Proof.** Let $\tau$ be the chain component of $E$ containing $\{h|(h,\{a,b\}) \in W\}$. Once again, $O \subseteq W$ is valid if, and only if, there exists a perfect ordering $o$ of $E_\tau$ such that the removal from the perfect directed version $H$ of $E_\tau$ creates the v-structures of $O$ but not those of $W \setminus O$. One the one hand, if $C$ is complete, then $C \cup \{a, b\}$ is complete. By lemma 2.2, if $h_1, \ldots, h_k$ is a permutation of $C$ then there exists a perfect ordering $o$ beginning with $h_1, \ldots, h_k, a, b$. That $o$ has the required properties. On the other hand, let $o$ be such a perfect ordering. Suppose that $a \rightarrow b \in H^{13}$. For each $h \in C$, $H_{\{a,b,h\}}$ is the subgraph of figure 10(a) or 10(b). As can be seen, by combining

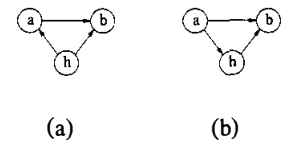

(a)          (b)

Figure 10: Induced Subgraphs of $E$

those subgraphs, every $h_i, h_j \in C$ must be adjacent in $H$.

---
[13]This is a matter of notation.

Otherwise $H$ would possess the v-structure $(b, \{h_i, h_j\})$. $C$ is thus complete in $E$. ∎

Given a $O$ characterizing $E' \in N_{ab}(E)$, $E'$ can be constructed with a procedure analogous to the one given in section 4.1. Let $G$ be the graph obtained from $E$ by removing $a - b$ and directing the lines of $E_\tau$ according to a perfect ordering of $E_\tau$ beginning with a permutation of $C$ followed by $a, b$. The arrows of $G$ present in $G_C$ are then undirected. One can see from the proof of theorem 4.12 that $G$ satisfies the hypotheses of theorem 2.10 and can thus be used as a starting point for algorithm 2.1. Besides, if $E$ does not induce a subgraph of the type of figure 9, i.e. $W = \emptyset$, then $E'$ can be constructed by a simpler procedure. Indeed, by theorem 2.7, the graph $G$ obtained by removing $a - b$ from $E$ is essential. Moreover, $G$ has the same skeleton and set of v-structures as $E'$. Hence, $E' = G$.

The increment in score is given by the next formula, where $a$ and $b$ can be permuted by symmetry.

$$\Delta_G score = f(b, pa_E(b) \cup C) - f(b, pa_E(b) \cup C \cup \{a\})$$

## 4.3 ADDITION OF AN EDGE TO $E$

The operation associated to $\{a, b\}$ is the addition of an arrow between $a$ and $b$. We use $c_1$. We have the following lemma.

**Lemma 4.13** *If $(h, \{t_1, t_2\}) \in R(\{a, b\})$, then (i) $\{t_1, t_2\} = \{a, t\}$, $h = b$ and $E$ induces the subgraph of figure 11(a) or 11(b), or (ii) $\{t_1, t_2\} = \{b, t\}$, $h = a$ and $E$ induces the subgraph of figure 11(c) or 11(d).*

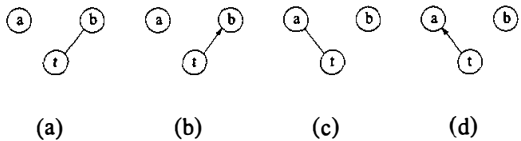

Figure 11: Induced Subgraphs of $E$

Let $P$ be the set of v-structures verifying the thesis of lemma 4.13. For each valid $(\{a, b\}, O)$, we have $O \subseteq R(\{a, b\}) \subseteq P$.

We didn't find *simple* graphical necessary and sufficient constraints on $E$ to determine the validity of a given characterization $O$, but we have lemma 4.14 and theorem 4.15. Let us introduce some notation. Let $P_i, i = 1, \ldots, 4$ be the partition of $P$ such that, for each element of $P_1, P_2, P_3$ or $P_4$, $E$ induces a subgraph of the type of, respectively, figure 11(a), 11(b), 11(c) or 11(d). Each valid $O \subseteq P$ can be decomposed into the sets $O_i = O \cap P_i, i = 1, \ldots, 4$.

**Lemma 4.14** *If $O$ is valid, then $O \subseteq P$ and (at least) one of the two following conditions is satisfied.*

(i) $O_2 = P_2$, $O_3 = O_4 = \emptyset$ and $F_1 = \{t|(b, \{t, a\}) \in O_1\}$ *is complete in $E$;*

(ii) $O_4 = P_4$, $O_1 = O_2 = \emptyset$ and $F_3 = \{t|(a, \{t, b\}) \in O_3\}$ *is complete in $E$.*

**Proof.** By lemma 4.3, $\exists K, L \in \mathcal{D} : L^* = E, K$ is obtained by adding to $L$ an arrow between $a$ and $b$, and $O = V(K) \setminus V(L)$. Let $\tau$ be the chain component of $E$ containing the set of vertices $\{t|(b, \{t, a\}) \in P_1\}$. By theorem 2.11, the arrows of $L_\tau$ are oriented according to a perfect ordering of $E_\tau$. Moreover, $t \to b \in L_\tau$ for $t \in F_1$. Every $t_i, t_j$ are adjacent in $L_\tau$, because otherwise $L_\tau$ would possess the v-structure $(b, \{t_i, t_j\})$. $F_1$ is thus complete in $E$. Similarly, we deduce that $F_3$ is complete in $E$. If $a \to b \in K$, then, by lemma 4.6, $O_2 = P_2$ and $O_3 = O_4 = \emptyset$. If $b \to a \in K$, then $O_4 = P_4$ and $O_1 = O_2 = \emptyset$. ∎

Suppose that a given $O$ satisfies these conditions. Let $G(O)$ be the graph obtained from $E$ as follows. If $O = \emptyset$, simply add $a - b$. Otherwise[14], if (i) is satisfied, add $a \to b$ and direct every line $t - b$ such that $t \in F_1$ towards $b$, while if (ii) is satisfied, add $b \to a$ and direct every line $t - a$ such that $t \in F_3$ towards $a$. We can check the validity of $O$ with the next theorem and Dor and Tarsi's algorithm.

**Theorem 4.15** *$O$ is valid if, and only if, $O$ satisfies the conditions of lemma 4.14 and $G(O)$ has a consistent extension.*

**Proof.** Suppose $G(O)$ has a consistent extension $M$. The essential graph $M^*$ is characterized by $(\{a, b\}, O)$. Indeed, $S(M^*) = S(G)$ and $V(M^*) \setminus V(E) = V(M) \setminus V(E) = V(G) \setminus V(E)$. By construction, $S(G)$ is characterized by $\{a, b\}$ and $V(G) \setminus V(E) = O$. Suppose that $O$ characterizes $E' \in N_{ab}(E)$. Let $K$ be one DAG whose existence is mentioned in lemma 4.3. $K$ is a consistent extension of $G(O)$. ∎

As this proof shows, given a $O$ characterizing $E' \in N_{ab}(E)$, $E'$ can be obtained by applying algorithm 2.1 to a consistent extension $M$ of $G(O)$[15]. $M$ can also be used to evaluate the increment in score.

## 5 APPLICATION TO LEARNING

In this section, the hypothetical use of our search space with greedy hill-climbing is discussed. This space has valuable properties. First, it is connected. Moreover, the score of each neighbour $E'$ of $E$ can be evaluated incrementally from $E$'s score and without constructing $E'$. If we do need $E'$ and $E' \in N^+(E)$, then it can be built from $E$ incrementally by retaining a priori some of its lines.

---
[14] The conditions (i) and (ii) of theorem 4.14 are now exclusive.
[15] The global nature of the acyclicity constraint prevents the incremental construction of the essential graphs with the previous procedure.

The main drawback of this search space is that the size of the neighbourhood can be intractable for structurally complex essential graphs. Indeed, let $c$ be the number of vertices of the largest complete undirected induced subgraph of $E$. Sections 4.1 to 4.3 tell us that, in the worst case, the number of elements of $N(E)$ is exponential in $c$. Let us make some early comments on the impact of this size on two opposite ways of starting a greedy hill-climbing search. Suppose that the search starts with the empty essential graph and then adds edges. We expect that the mean size of $N(E)$ and thus the computational cost will augment as we progress in the space. This behaviour is certainly problematic, but probably comes with a growing need for more data to support the successive removal of the independences. Suppose now that the search starts with the complete essential graph and then prunes it. In that case, our neighbourhood is clearly inappropriate. This can be interpreted as the fact that it is too fine-grained for pruning, at least in the early steps, and that a more aggressive strategy should be used.

## 6 CONCLUSION

The topic of this paper is the construction and analysis of a search space of Markov equivalence classes of Bayesian networks represented by essential graphs and with the inclusion boundary neighbourhood. Our analysis shows that this space is connected and the score of each neighbour of an equivalence class can be evaluated incrementally from the score of that class. Another important contribution is the suggestion of a procedure to actually build the neighbourhood of a class. As a byproduct, a bound on the size of the neighbourhood that can be calculated very simply a priori is determined.

This work can be extended by a careful estimation of the impact of that size on the learning algorithms to possibly propose approximations. In a next step, this space can be compared to others, based on Bayesian networks or equivalence classes, for example on the basis of the performance of the algorithms using them.